%% file: arxiv.tex
\documentclass{article}

\usepackage[preprint,nonatbib]{neurips_2025}

\usepackage[utf8]{inputenc} 
\usepackage[T1]{fontenc}    
\usepackage{hyperref}       
\usepackage{url}            
\usepackage{booktabs}       
\usepackage{amsfonts}       
\usepackage{nicefrac}       
\usepackage{microtype}      
\usepackage{xcolor}         

\usepackage{graphicx}
\usepackage{amsmath}
\usepackage{amssymb}
\usepackage{multirow}
\usepackage{pifont}    
\usepackage[normalem]{ulem}
\useunder{\uline}{\ul}{}
\usepackage[inline]{enumitem}
\usepackage{enumitem}
\usepackage{wrapfig} 

\title{Reinforced Interactive Continual Learning via Real-time Noisy Human Feedback}

\author {
    Yutao Yang\textsuperscript{\rm 1},
    Jie Zhou\textsuperscript{\rm 1}\thanks{Corresponding author, jzhou@cs.ecnu.edu.cn.} , 
    Junsong Li\textsuperscript{\rm 1}, 
    Qianjun Pan\textsuperscript{\rm 1}, 
    Bihao Zhan\textsuperscript{\rm 1},\\
    \textbf{Qin Chen}\textsuperscript{\rm 1}, 
    \textbf{Xipeng Qiu} \textsuperscript{\rm 2,3}, 
    \textbf{Liang He}\textsuperscript{\rm 1} \\
    \textsuperscript{\rm 1}School of Computer Science and Technology, East China Normal University \\
    \textsuperscript{\rm 2}Shanghai Innovation Institute 
    \textsuperscript{\rm 3}School of Computer Science, Fudan University. 
}

\begin{document}

\maketitle

\begin{abstract}
This paper introduces an \textbf{interactive continual learning} paradigm where AI models dynamically learn new skills from \textbf{real-time human feedback} while retaining prior knowledge. This paradigm distinctively addresses two major limitations of traditional continual learning: (1) dynamic model updates using streaming, real-time human-annotated data, rather than static datasets with fixed labels, and (2) the assumption of clean labels, by explicitly handling the noisy feedback common in real-world interactions. To tackle these problems, we propose \texttt{RiCL}, a Reinforced interactive Continual Learning framework leveraging Large Language Models (LLMs) to learn new skills effectively from dynamic feedback. \texttt{RiCL} incorporates three key components: a temporal consistency-aware purifier to automatically discern clean from noisy samples in data streams; an interaction-aware direct preference optimization strategy to align model behavior with human intent by reconciling AI-generated and human-provided feedback; and a noise-resistant contrastive learning module that captures robust representations by exploiting inherent data relationships, thus avoiding reliance on potentially unreliable labels. Extensive experiments on two benchmark datasets (FewRel and Tacred), contaminated with realistic noise patterns, demonstrate that our \texttt{RiCL} approach substantially outperforms existing combinations of state-of-the-art online continual learning and noisy-label learning methods.
\end{abstract}

\input{section/introduction}

\input{section/related_work}

\input{section/method}

\input{section/experiment}

\section{Conclusions and Further Work}
To learn like a human, we introduce a Reinforced interactive Continual Learning (\texttt{RiCL}) framework to learn in real time using the noisy feedback from human-machine interaction. 
Particularly, we design a reinforcement learning based method to improve the model's performance via the gap between the labels given by AI and humans. 
Extensive experimental results on two benchmarks denote that \texttt{RiCL} consistently surpasses strong online CL and noisy-label learning baselines, validating its robustness to both label noise and task-order variation. Ablation studies corroborate the contribution of each component, including the Temporal Consistency-aware Purifier (TCP), Noise-resistant Contrastive Learning (NCL), and Interaction-aware Preference Optimization (IPO). 

Despite its strengths, \texttt{RiCL} has a few limitations. First, its purifier is tuned for symmetric noise, which may not fully handle real-world feedback with biases, adversarial inputs, or context-dependent errors. Second, it only supports single-turn feedback, missing multi-turn interactions, delayed responses, or partial credit — key for complex human interactions. Future work could explore scaling to larger models, adding multimodal feedback, and developing richer evaluation metrics for lifelong learning.


\bibliographystyle{unsrt}  

\bibliography{references}  


\appendix

\input{section/appendix}

\end{document}

%% file: section/introduction.tex
\section{Introduction}  

\begin{wrapfigure}{r}{0.6\textwidth}
  \vspace{-3mm}  
  \centering
  \includegraphics[width=0.58\textwidth]{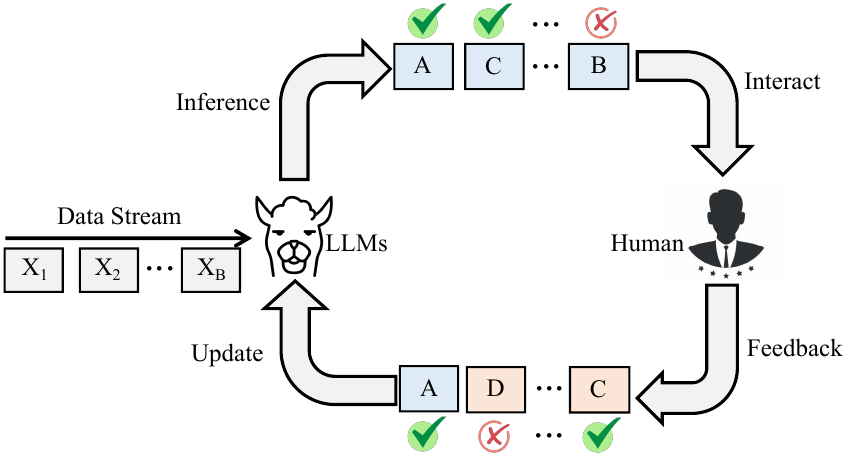}
  \vspace{-2mm}
  \caption{The process of Interactive Continual Learning.}
  \label{fig:intro}
  \vspace{-1mm}
\end{wrapfigure}

Continual learning (CL) has emerged as a critical paradigm for enabling artificial intelligence (AI) systems to continuously adapt and evolve in dynamic environments \cite{wang2024comprehensive}. 
Unlike traditional machine learning approaches, which often rely on static datasets and fixed task boundaries, continual learning emphasizes incremental knowledge acquisition while preserving previously learned capabilities \cite{li2017learning,kirkpatrick2017overcoming,de2021continual}. Recent advancements in AI, particularly large language models (LLMs), have further highlighted the potential of interactive and adaptive systems that learn through real-world interactions \cite{shaheen2022continual,liu2021lifelong,yang2025recent}. 
For real-world interactive scenarios (e.g., personalized assistance, autonomous systems, or adaptive decision-making platforms), the model outputs the predicted labels to the human, and the human gives feedback to update the model (as shown in Figure \ref{fig:intro}).

Most traditional continual learning approaches often rely on pre-collected static datasets with given labels \cite{huai2025cl}. They mainly focus on reducing the catastrophic forgetting problem via replay-based \cite{rolnick2019experience}, regularization-based \cite{ahn2019uncertainty}, and architecture-based \cite{ding2024boosting} methods. 
Recently, some studies have focused on online continual learning that learns new knowledge with data streaming to adapt in real-time \cite{bidaki2025onlinecontinuallearningsystematic,mai2022online,wang2023cba}. 
Qi et al. \cite{qi2024interactive} introduced an interactive continual learning framework that enables collaborative interactions among multiple models of varying sizes.
However, few studies have explored \textbf{continuous learning from }\textbf{human-machine interaction}, with a focus on dynamic model adaptation based on noisy human feedback.

These methods face two significant challenges prevalent in real-world interactive scenarios: (1) Evolving over time via the data stream from real-time human feedback. As illustrated in Figure \ref{fig:intro}, this task involves feeding a continuous data stream into LLMs while incorporating real-time human feedback to refine the model's performance without forgetting the previous abilities; (2) Robustly learning from inherently noisy human feedback, which is a common characteristic of interactive applications. In these scenarios, humans provide corrections to the predictions made by LLMs, which may sometimes be incorrect. Furthermore, existing studies often overlook the importance of leveraging the predicted results for further improvement. These limitations impede the effective deployment of AI systems in environments where human-in-the-loop interactions and dynamic, real-time data updates are critical.

To bridge this gap, we propose \texttt{RiCL}, a Reinforced interactive Continual Learning framework grounded in LLMs, designed to harmonize real-time human feedback with stable knowledge retention. 
First, we design a temporal consistency-aware purifier (TCP), which dynamically discriminates between clean and noisy samples in streaming data by analyzing temporal coherence and prediction stability, enabling selective integration of reliable information. Then, we propose an interaction-aware direct preference optimization (IPO), which aligns model behavior with human intent by resolving conflicts between AI-generated outputs and human feedback, leveraging a reinforcement learning paradigm to prioritize feedback that enhances task coherence and user satisfaction. 
Finally, we introduce a noise-resistant contrastive learning module that extracts robust feature representations by exploiting inherent semantic relationships in the data, reducing reliance on potentially noisy labels through self-supervised alignment of augmented samples.  
Our experiments on two datasets (Fewrel and Tacred) contaminated with real-world noise patterns demonstrate that \texttt{RiCL} outperforms state-of-the-art combinations of online CL and noisy-label learning methods. By bridging the gap between theoretical lifelong learning principles and practical interactive AI systems, this work advances the development of autonomous, adaptive models capable of thriving in open-world environments.
In summary, our primary contributions are threefold:
\begin{itemize}[leftmargin=*, align=left]
    \item To learn like a human, we propose a reinforced interactive continual learning framework to learn new knowledge without forgetting previous skills in real time via noisy human feedback from human-machine interaction. 
    \item We design an interaction-aware direct preference optimization and a noise-resistant contrastive learning strategy to learn continually from clean and noisy datasets that are indicated by a temporal consistency-aware purifier. 
    \item Our extensive experiments demonstrate that \texttt{RiCL} obtains better performance in both mitigating catastrophic forgetting and handling label noise, offering actionable insights for robust continual learning in real-world streaming interactive scenarios.
\end{itemize}

%% file: section/related_work.tex
\section{Related Work}
\paragraph{Continual Learning}
Continual learning enables artificial intelligence models to incrementally learn new tasks without forgetting previously acquired knowledge \cite{yang2025recent}.
The literature on continual learning predominantly categorizes methods into three classes: rehearsal-based \cite{rolnick2019experience}, regularization-based \cite{wang2023orthogonal}, and parameter-isolation \cite{ding2024boosting} approaches. 
Rehearsal-based methods preserve previously learned knowledge by retaining subsets of past samples or generating pseudo-samples (generative replay) to periodically revisit during new task learning, mitigating catastrophic forgetting \cite{shin2017continual,van2020brain,du2023static}. 
For example, LAMOL \cite{sun2019lamol} generates pseudo-samples to rehearse past NLP tasks, and RVAE\_LAMOL \cite{wang2022rvae} utilizes a variational autoencoder for stable and accurate pseudo-sample generation. 
Regularization-based approaches impose constraints or penalties on parameter updates to balance new learning and knowledge retention, preventing significant deviations from previously learned representations \cite{huang2021continual}. For instance, Continual Proximal Policy Optimization (CPPO) \cite{zhang2024cppo} integrates sample-wise weighting into policy optimization to maintain prior knowledge while learning new policies effectively. 
Parameter-isolation methods allocate separate parameters or subnetworks for distinct tasks, explicitly avoiding interference among tasks \cite{gao2023unified}. JARe \cite{peng2024scalable} dynamically utilizes task-related knowledge retrieval to effectively manage parameter adjustments for different downstream tasks.

Notably, Qi et al. \cite{qi2024interactive} proposed Interactive Continual Learning (ICL), which leverages interactions between different model complexities to enhance memory retrieval and collaborative reasoning capabilities. 
Moreover, Xu et al. \cite{xu2018reinforced} utilized reinforcement learning to dynamically optimize neural architectures for continual learning. 
Unlike these studies, we focus on an interactive continual learning task to learn real-time human feedback from interactions, prioritizing alignment between human intent and model output.

\paragraph{Online Continual Learning}

Online continual learning explores scenarios where models continuously adapt to new knowledge by sequentially or concurrently processing incoming data streams \cite{aljundi2019gradient,aljundi2019online}. 
Studies increasingly focus on effective knowledge retention and adaptation in realistic, dynamically evolving scenarios \cite{liu2020learning}. 
Current research categorizes it into two main settings based on task boundary clarity: hard task boundary \cite{qiao2024gradient} and blurry task boundary \cite{bang2022online,aljundi2019task}.

In a hard task boundary configuration, tasks arrive sequentially and are distinctly structured. Each task is fully processed before transitioning to the subsequent one, ensuring a clear separation of data across tasks. Representative methods include episodic memory-based techniques, such as MBPA++ \cite{de2019episodic}, which utilize episodic replay for knowledge retention, and gradient-based approaches like PEGP \cite{qiao2024gradient}, which employ parameter-efficient gradient projections to alleviate catastrophic forgetting.
Conversely, the blurry task boundary configuration closely resembles real-world scenarios where task distinctions become ambiguous or overlap significantly. Here, data streams from multiple tasks are intermixed, complicating the clear identification of task transitions. Methods developed to address such conditions include SIT \cite{wang2024clip}, which employs contrastive learning to implicitly differentiate tasks, and G-NoCL \cite{seo2024just}, which dynamically adapts learning objectives to effectively manage overlapping data distributions.
Most existing studies are based on the assumption that the labels of the samples are provided. However, the human feedback from the data stream in real-world interaction is not well studied, which is essential for machine learning.

\paragraph{Noisy Label Learning}
In parallel to continual learning, addressing the challenge of noisy labels has also drawn sustained research interest. Strategies in this area include model regularization \cite{wang2019symmetric, wei2020combating}, label correction \cite{yi2019probabilistic, song2019selfie}, and sample filtering \cite{han2018co, yu2019does}. Among these, sample filtering strategies—particularly training models primarily on samples with small-loss values presumed reliable—are widely adopted. MentorNet \cite{jiang2018mentornet} employs a teacher network guiding the student network by selecting reliable small-loss samples, while Co-teaching \cite{han2018co} and Co-teaching+ \cite{yu2019does} use dual-model structures to mutually identify and exchange such trustworthy samples. Label correction methods, such as SELFIE \cite{song2019selfie}, adjust noisy labels based on consistency checks, whereas SEAL \cite{chen2021beyond} refines labels through averaged softmax outputs across training epochs. 
These methods, although effective in static scenarios, typically assume the availability of pre-prepared labeled data, highlighting the need for novel techniques to handle noisy labels effectively within dynamic interactions.

\begin{figure}[t]
\centering
\includegraphics[width=1.0\columnwidth]{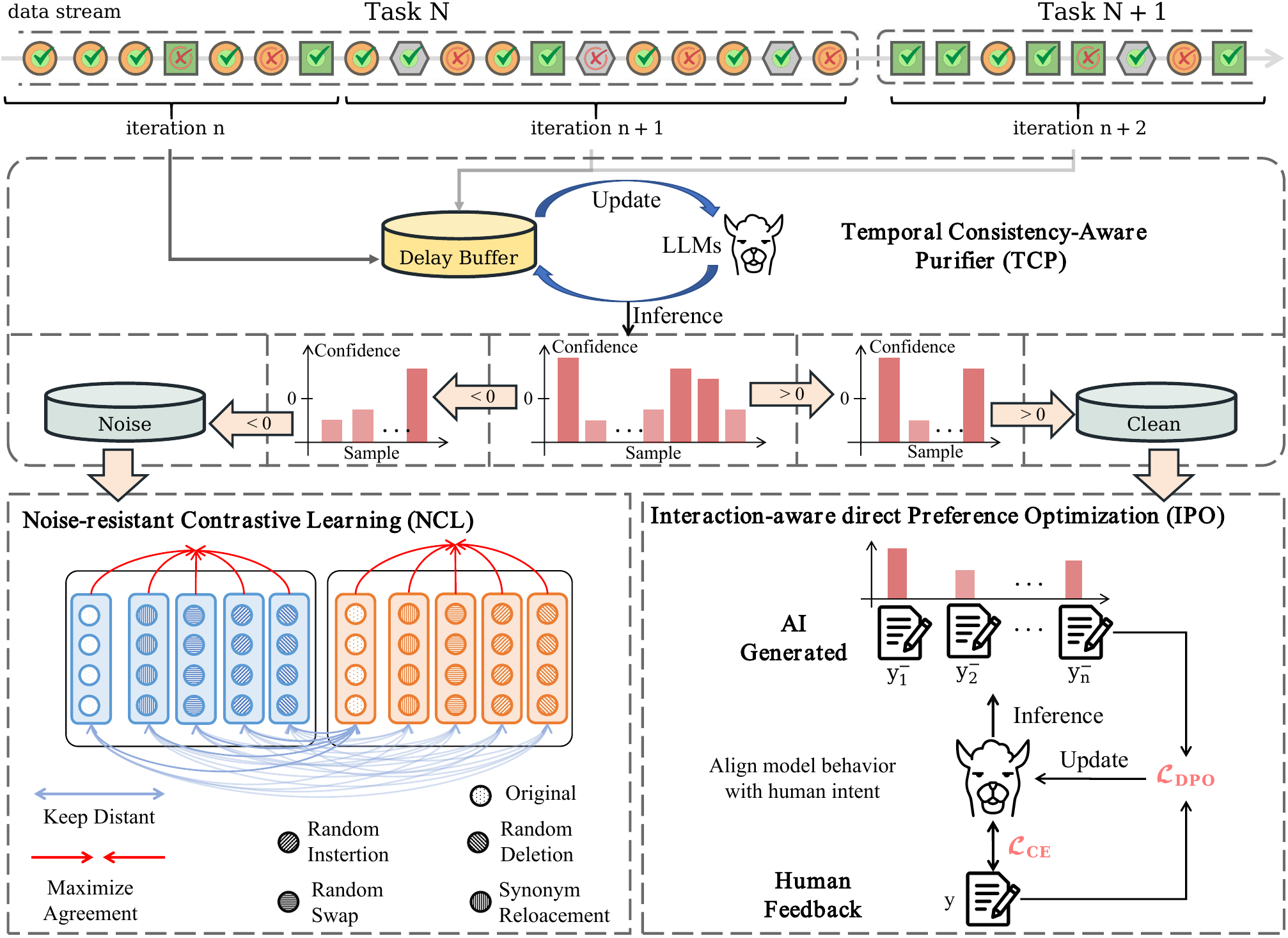}
\vspace{-1mm}
\caption{The framework of our Reinforced interactive Continual Learning (\texttt{RiCL}).}
\label{fig:framework}
\vspace{-2mm}
\end{figure}

%% file: section/method.tex
\section{Methodology}
In this paper, we propose a Reinforced interactive Continual Learning (\texttt{RiCL}) framework to learn knowledge from human feedback (Figure \ref{fig:framework}). It consists of three parts: temporal consistency-aware purifier, interaction-aware direct preference optimization, and noise-resistant contrastive learning.
First, we classify the streaming data into clean and noisy datasets via a temporal consistency-aware purifier. Then, we introduce an interaction-aware direct preference optimization to learn from human feedback by reducing the gap between the results predicted by models and the answers corrected by humans. 
Finally, we propose a noise-resistant contrastive learning module to learn a robust text representation from noisy data via data argumentation.

\subsection{Task Formulation}


In this work, we address a human-in-the-loop continual learning paradigm characterized by streaming data and iterative human-model interaction. The learning process involves sequential tasks where the model's predictions are continuously refined through human feedback. Formally, we train our model on the sequence of tasks $T=\{T_1, ..., T_{N}, T_{N+1}, ... T_{|T|}\}$. For each task $T_N$, the data is coming in a stream. We use a delay buffer to stock the stream data and update our models for each buffer, where the number of samples is $M$ in the $n$-th delay buffer ${D_n}=\{X_i = (x_i, \tilde{y_i}, y_i)\}^{M}_{i=1}$. For each sample $X=(x, \tilde{y}, y)$, $x$ denotes input instances, $\tilde{y} \in C$ represents model-generated labels from the label space $C$, and $y \in C$ corresponds to potentially noisy human-provided corrections.

\subsection{Temporal Consistency-aware Purifier}
In streaming data scenarios, the Temporal Consistency-aware Purifier (TCP) distinguishes clean samples from noisy ones by examining the stability of predictions and temporal coherence. It tracks whether model outputs remain consistent over time for the same or similar data points, leveraging that consistency to assess data reliability. 

First, to obtain a noise-robust purifier, we train the model \(\mathcal{M}_p^n\) at the \(n\)-th delay buffer using the Generalized Cross-Entropy (GCE) loss \cite{zhang2018generalized}. It reduces sensitivity to noisy labels by interpolating between traditional cross-entropy and mean absolute error losses. 
By penalizing large deviations while incorporating previous training outcomes, GCE effectively balances robustness and precision through this iterative update scheme. 
Formally, the GCE loss for the \(n\)-th delay buffer is written as:
\begin{equation}
    \mathcal{L}_{\text{GCE}} 
    = - \frac{1}{M} 
    \sum_{(x, \tilde{y}, y) \in D_n}
    \log \biggl( 
      1 + \exp( -\,y \cdot \mathcal{M}_p^n(x; \theta_p^{n}) )
    \biggr)
\end{equation}
where \(\mathcal{M}_p^n(x;\theta_p^{n})\) represents the purifier model’s prediction at the \(n\)-th delay buffer, whose parameters \(\theta_p^{n}\) are initialized from \(\theta_p^{n-1}\) of \(\mathcal{M}_p^{n-1}\). 




After the purifier is trained, we calculate the temporal confidence consistency based on the average uncertainty measure \cite{pleiss2020identifying} to evaluate the ``cleanliness" of each sample in the streaming data. 
For each sample, we calculate the confidence based on the insight that correctly labeled instances progressively converge towards their labels, manifesting positive logit margins. Conversely, mislabeled samples exhibit persistently low or negative margins.
\begin{equation}
    \text{Confidence}(x) = \text{logit}(y) - \max_{c \neq y}\text{logit}(c)
\end{equation}
where $\text{logit}=\mathcal{M}_p^n(x; \theta^n_{p})$, \(\text{logit}(y)\) indicates the predicted logit of the labeled class, and \(\max_{c \neq y}\text{logit}(c)\) represents the highest logit among other classes. 
For each sample \( X \), the purifier $\mathcal{M}_p^n$ first computes its $\text{Confidence}(x)$ score to decide whether to place it into the clean delay buffer ($\mathcal{C}$) or noisy delay buffer ($\mathcal{N}$). 
\begin{equation}
    x \mapsto
\begin{cases}
\mathcal{C}, & \text{if } \text{Confidence}(x)>=0, \\
\mathcal{N}, & \text{if } \text{Confidence}(x)<0.
\end{cases}
\end{equation}
Once the delay buffer reaches capacity, the latest purifier $\mathcal{M}_p^{\text{new}}$ recalculates the $\text{Confidence}^{\text{new}}(x)$ for all buffered samples to assess temporal prediction consistency. If the two confidence results for a sample remain consistent, that sample is then moved into the corresponding replay buffer ($\mathcal{R}_{\text{clean}}$ or $\mathcal{R}_{\text{noisy}}$). 
Particularly, we compute the temporal confidence consistency as follows:
\begin{equation}
       x \mapsto
   \begin{cases}
   \mathcal{R}_{\text{clean}}, & \text{if } \text{Confidence}(x) 
      = \text{Confidence}^{\text{new}}(x)>=0,\\[6pt]
   \mathcal{R}_{\text{noisy}}, & \text{if } \text{Confidence}(x) 
      = \text{Confidence}^{\text{new}}(x)<0,\\[6pt]
   \varnothing, & \text{otherwise (discard)}.
   \end{cases}
\end{equation}


\subsection{Interaction-aware Direct Preference Optimization}
In this section, we propose an Interaction-aware Direct Preference Optimization (IPO) to align the model with human feedback. 
Specifically, we train the primary LLM \(\mathcal{M}^n\) using reinforcement learning to exploit the gap between the model-generated labels and human feedback. 
We construct preference data based on this gap to train \(\mathcal{M}^n\) via Direct Preference Optimization (DPO) \cite{rafailov2023direct}.
To reduce the catastrophic forgetting problem, we train \(\mathcal{M}^n\) on both the clean samples identified by TCP from the current data stream $D_n$ and the clean data stored in the clean replay buffer ($\mathcal{R}_{\text{clean}}$). 

The preference dataset for input sample $x$ is $\mathcal{P} = \left\{(y, \tilde{y}^{(j)}) \mid j \in [1, L]\right\}$, where $\tilde{y}^{(j)}$ is the $j$-th alternative predicted label. Instead of choosing the top \( L \) highest-probability incorrect labels, alternative labels are sampled proportionally from the model’s predicted distribution, explicitly excluding the correct label. 
Each alternative label \(\tilde{y}^{(j)}\) is sampled according to a categorical distribution derived from the softmax-normalized logits over incorrect labels:
\begin{equation}
    \mathbb{P}(\tilde{y}^{(j)} = y') = \frac{\exp(z_{y'})}{\sum_{y'' \ne y}\exp(z_{y''})}, \quad y' \ne y
\end{equation}
Here, \( z_{y'} \) represents the model \(\mathcal{M}_p^n(x;\theta_p^{n})\)'s logit for class \( y' \). 
This sampling strategy ensures that alternative labels are selected proportionally to their predicted likelihood, thus providing meaningful contrasts for the preference pairs. 

Each preference pair trains the model explicitly to differentiate the correct label from plausible alternatives, enhancing decision-making robustness. The preference-based loss function is:
\begin{equation}
    \mathcal{L}_{\text{IPO}} = -\sum_{j=1}^{L} \log \sigma\left(\log p^n_{\theta}(y|x) - \log p^n_\theta(\tilde{y}^{(j)}|x)\right)
\end{equation}
Here, \( p^n_\theta(\cdot) \) is the conditional probability predicted by the model \(\mathcal{M}^n\) that is initialed by \(\mathcal{M}^{n-1}\), and \(\sigma(\cdot)\) represents the sigmoid function. By maximizing the logit difference between original and alternative labels, the model learns robust decision boundaries, prioritizing true labels. 
Note that we train \(\mathcal{M}^n\) through supervised fine-tuning before reinforcement learning to improve the stability.

\subsection{Noise-resistant Contrastive Learning}
Furthermore, we integrate Noise-resistant Contrastive Learning (NCL) to learn a robust feature representation using the noisy data that is filtered by TCP from the current data stream and stored in the noisy replay buffer ($\mathcal{R}_{\text{noisy}}$). 
Specifically, we encourage the primary model to capture intrinsic semantic structures using these noise-identified samples with contrastive learning.

The contrastive loss is computed between each original sample and its $k$ augmented variants \(\{x_j^+\}_{j=1}^k\), thereby expanding the diversity of the training dataset. Formally, the total contrastive loss for each augmented pair is defined as:
\begin{equation}
   \mathcal{L}_{\text{NCL}} = - \log \,
  \frac{ \displaystyle \sum\nolimits_{j=1}^{k} \exp(\mathrm{Score}(x,x_j^+)\,/\,\tau)}
       { \displaystyle \sum\nolimits_{z \in \mathcal{B} \land z \neq x} \exp(\mathrm{Score}(x, z)\,/\,\tau ) + \sum\nolimits_{j=1}^{k} \exp(\mathrm{Score}(x, z^{+}_{j})\,/\,\tau)}
\end{equation}

where $\mathrm{Score}(a,b)=\exp(\mathrm{sim}(f_\theta(a),\,f_\theta(b))$, \( f_\theta(\cdot) \) is the embedding representation from model \(\mathcal{M}_\theta\) and $\mathrm{sim(\cdot,\cdot)}$ calculates the cosine similarity, and batch \( \mathcal{B} \) encompasses all samples along with their augmented versions.

We employ four data augmentation methods ($k$ = 4)  to enhance data diversity: 1) Synonym replacement replaces non-stopwords with synonyms, preserving meaning while introducing lexical variation; 2) Random insertion inserts synonyms of randomly selected words into new positions, enriching contextual representation; 3) Random swap exchanges the positions of two random tokens, perturbing syntax without significantly altering semantics; and 4) Random deletion removes words with a fixed probability, encouraging robustness by simulating partial input scenarios. These techniques collectively promote semantic and syntactic diversity, improving model generalization.



The overall training objective of the main model integrates multiple optimization stages designed to effectively address noisy labels and leverage clean samples. Initially, the model is trained using noise-resistant contrastive learning on samples identified as noisy by the TCP. Subsequently, IPO is applied exclusively to clean samples. 

%% file: section/experiment.tex
\section{Experimental Setups}

\paragraph{Datasets}
Due to the lack of datasets for interactive continual learning, we evaluate \texttt{RiCL} using two simulated datasets: Fewrel \cite{han-etal-2018-Fewrel} and Tacred \cite{zhang-etal-2017-position}. We add noise to the ground truth and regard these labels as human feedback. 
For Fewrel, we generate a sequence of 10 tasks, each comprising eight relation classes selected randomly without replacement, as in \cite{li2023online}. Similarly, we partition Tacred into 10 sequential tasks, each containing four distinct relation classes. 

We utilize the Blurry-CL scenario \cite{bang2022online} to incorporate class overlap among sequential tasks intentionally. 
Specifically, each class is designated as a primary class exactly once, while serving as a secondary class in all other tasks.
We design a blur rate $r = \frac{|\bigcup_{c \in m_t} \mathcal{D}_{t,c}|}{|\mathcal{D}_t|}$ as a overlap parameter, where \( \mathcal{D}_t \) denotes the dataset of task \( t \) and \( \mathcal{D}_{t,c} \) represents instances of class \( c \). In our experiments, we set \( r\) to 0.1, indicating that secondary classes constitute roughly 10\% of the classes within each task. 

To realistically simulate label corruption scenarios, we introduce symmetric label noise \cite{han2018co, li2020dividemix}. This approach ensures each true label has an equal likelihood of being erroneously reassigned to any other label within the same task, thereby rigorously testing the model's robustness to mislabeled data.
This systematic arrangement enables the simulation of realistic continual learning scenarios.

\paragraph{Evaluation Metrics}
Following \cite{chaudhry2018riemannian, lopez2017gradient}, we employ two established metrics, final Average Performance (AP) and Average Forgotten (AF) to evaluate the performance. 
AP measures the model's capability to retain previously acquired knowledge when sequentially trained on new tasks, calculated as $AP = \frac{1}{M} \sum_{i=1}^{M} m_{M, i}$, where \( m_{M, i} \) represents the accuracy on task \( i \) following training completion on all \( M \) tasks. AF quantifies the extent of forgetting, computed as $AF = \frac{1}{M-1} \sum_{i=1}^{M-1}(m_{i, i} - m_{M, i})$, where \( m_{i,i} \) is the immediate performance on task \( i \) after its initial training, and \( m_{M,i} \) denotes performance on task \( i \) after all subsequent training.

\paragraph{Baseline Methods}
To comprehensively evaluate the effectiveness of our method, we compare it against several established baseline approaches. First, we consider five widely recognized online continual learning techniques: ER \cite{rolnick2019experience}, MIR \cite{aljundi2019online}, and I-LoRA \cite{li2025analyzing}, SSR \cite{huang2024mitigating}, along with an online noisy-label learning approach, S6 \cite{li2023online}. Additionally, we combine the strong ones with two robust noisy label learning strategies, namely SL \cite{wang2019symmetric} and JoCoR \cite{wei2020combating}. For further reference, we also employ two control setups: Multitask training is always regarded as the upper bound, which simultaneously learns all tasks, and Sequential Finetuning (SeqFT), which sequentially learns each task without applying explicit noise mitigation techniques.

\paragraph{Implementation Details}
In our experiments, all continual learning methods employ LLaMA-7B \cite{touvron2023llama} as the backbone model to maintain consistency and fairness across comparisons. We set the replay buffer size to 4,000 instances for Fewrel and 800 instances for Tacred. 
Specifically, for Fewrel, \texttt{RiCL}'s dataset partition sizes are set as follows: \(|\mathcal{D}|=1,000\), \(|\mathcal{C}|=1,000\), and \(|\mathcal{N}|=2,000\). Correspondingly, for Tacred, these sizes are adjusted to \(|\mathcal{D}|=200\), \(|\mathcal{C}|=200\), and \(|\mathcal{N}|=400\). All experiments are run on two NVIDIA A800 GPUs ($\approx$ 40 GB VRAM each), with each run taking roughly 30 hours.
For noisy-label learning methods, hyperparameters are configured as follows: SL adopts parameters \(\alpha = \beta = 1.0\), while JoCoR uses \(\lambda = 0.1\). To further ensure fairness in the comparative evaluation, we align \texttt{RiCL}'s replay buffer size closely with these configurations.

\begin{table*}[!t]
\centering
\small
\setlength{\tabcolsep}{0.6mm}{
\begin{tabular}{lcccccccccccc}
\specialrule{1.5pt}{0pt}{0pt}
\multirow{2}{*}{Methods} & \multicolumn{10}{c}{Various task in Tacred}                                                                                                                             & \multirow{2}{*}{$AP(\uparrow)$} & \multicolumn{1}{l}{\multirow{2}{*}{$AF(\downarrow)$}} \\ \cline{2-11}
                         & Task 1           & Task 2           & Task 3           & Task 4           & Task 5           & Task 6           & Task 7           & Task 8           & Task 9           & Task 10           &                     & \multicolumn{1}{l}{}                     \\ 
\hline
SeqFT & 55.90 & 66.43 & 67.88 & 46.63 & 85.11 & 58.22 & 51.09 & 69.83 & 51.65 & 94.20 & 64.69 & 30.44 \\ 
MIR & 61.49 & 61.54 & 73.72 & 50.92 & 82.98 & 61.64 & 60.87 & 56.03 & 64.84 & 85.51 & 65.95 & 18.90 \\
ER & 62.11 & 72.73 & 72.26 & 49.08 & 80.14 & 67.12 & 63.04 & 55.17 & 68.13 & 73.91 & 66.37 & 16.93\\
SSR  & 24.22 & 67.83 & 54.74 & 49.08 & 80.85 & 52.74 & 69.57 & 65.52 & 76.92 & 91.30 & 63.28 & 14.10 \\
S6 & 90.68 & 70.63 & 75.91 & 61.96 & 87.94 & 57.53 & 57.61 & 68.97 & 80.22 & 53.62 & 70.51 & -\\ 
I-LoRA & 75.16 & 80.42 & 79.56 & 70.55 & 80.85 & 68.49 & 70.65 & 58.62 & 76.92 & 78.26 & 73.95 & 8.26 \\
\hline
SeqFT+SL & 62.73 & 67.83 & 61.31 & 46.01 & 83.69 & 53.42 & 60.87 & 69.83 & 62.64 & 92.75 & 66.11 & 31.62 \\ 
MIR+SL & 60.25 & 69.23 & 72.99 & 55.83 & 78.01 & 65.75 & 58.70 & 47.41 & 54.95 & 86.96 & 65.01 & 20.53 \\
MIR+JoCoR & 66.46 & 72.03 & 56.93 & 46.63 & 85.82 & 58.22 & 72.83 & 63.79 & \textbf{83.52} & \textbf{94.20} & 70.04 & 19.61 \\
ER+SL & 72.05 & 70.63 & 74.45 & 49.08 & 76.60 & 62.33 & 61.96 & 56.90 & 60.44 & 81.16 & 66.56 & 18.79\\
ER+JoCoR & 75.78 & 76.92 & 75.91 & 55.83 & 80.85 & 71.92 & 70.65 & 64.66 & 70.33 & 86.96 & 72.98 & 15.00\\
I-LoRA+SL & 77.64 & 79.02 & 76.64 & 74.23 & 83.69 & 67.12 & 68.48 & 60.34 & 79.12 & 76.81 & 74.31 & 5.88\\
I-LoRA+JoCoR & 70.81 & 80.42 & 75.18 & 69.94 & 83.69 & 51.37 & 68.48 & 60.34 & 72.53 & 81.16 & 71.39 & 7.68 \\
\hline
\texttt{\textbf{RiCL}} & \textbf{85.09} & \textbf{90.21} & \textbf{84.67} & \textbf{80.37} & \textbf{87.94} & \textbf{78.84} & \textbf{77.41} & \textbf{78.92} & 78.02 & 85.51 & \textbf{82.70} & \textbf{1.69} \\ \hline
Multitask  & 89.44 & 91.61 & 81.75 & 77.30 & 90.78 & 77.40 & 81.52 & 71.55 & 92.31 & 82.61 & 83.63 & - \\
Multitask+SL & 88.20 & 93.01 & 81.75 & 80.37 & 92.2 & 78.08 & 81.52 & 74.14 & 91.21 & 79.71 & 84.02 & - \\
\specialrule{1.5pt}{0pt}{0pt}
\end{tabular}}
\caption{Performance (\%) of our \texttt{RiCL} and distinct continual learning method on Tacred. The best results are emphasized in \textbf{bold}. Multitask (+SL) is the upper bound.}
\label{tab_Tacred}
\vspace{-1mm}
\end{table*}

\begin{table*}[!t]
\centering
\small
\setlength{\tabcolsep}{0.6mm}{
\begin{tabular}{lcccccccccccc}
\specialrule{1.5pt}{0pt}{0pt}
\multirow{2}{*}{Methods} & \multicolumn{10}{c}{Various task in Fewrel}                                                                                                                             & \multirow{2}{*}{$AP(\uparrow)$} & \multicolumn{1}{l}{\multirow{2}{*}{$AF(\downarrow)$}} \\ \cline{2-11}
                         & Task 1           & Task 2           & Task 3           & Task 4           & Task 5           & Task 6           & Task 7           & Task 8           & Task 9           & Task 10           &                     & \multicolumn{1}{l}{}                     \\
\hline
SeqFT & 72.41 & 59.55 & 57.05 & 57.59 & 55.36 & 64.64 & 75.45 & 61.52 & 82.59 & 94.73 & 68.09 & 28.19 \\ %
MIR & 76.79 & 68.21 & 69.73 & 64.38 & 60.54 & 70.80 & 77.50 & 66.70 & 84.02 & 83.66 & 72.23 & 16.49 \\
ER & 78.66 & 66.25 & 71.61 & 67.59 & 67.86 & 74.02 & 79.20 & 69.82 & 81.07 & 87.86 & 74.39 & 15.70 \\
SSR & 85.23 & 78.31 & 72.44 & 66.02 & 64.73 & 68.54 & 70.05 & 61.91 & 65.82 & 70.57 & 70.36 & 13.20 \\
S6  & 87.68 & 73.93 & 84.82 & 82.5 & 73.04 & 75.8 & 63.12 & 50.45 & 67.23 & 63.57 & 72.21 & - \\ 
I-LoRA & 86.61 & 80.27 & 76.43 & 81.07 & 73.3 & 74.55 & 79.29 & 73.39 & 80.27 & 79.02 & 78.42 & 9.07 \\
\hline
SeqFT+SL & 76.79 & 62.23 & 59.46 & 60.0 & 56.52 & 69.20 & 75.36 & 64.29 & 85.00 & 92.50 & 70.14 & 26.43  \\ 
MIR+SL & 77.14 & 73.04 & 66.88 & 64.29 & 66.43 & 75.09 & 83.04 & 72.59 & 83.93 & 84.20 & 74.66 & 14.87 \\
MIR+JoCoR & 75.09 & 75.71 & 78.75 & 72.77 & 68.30 & 77.95 & 83.48 & 77.68 & 84.82 & 94.38 & 78.89 & 14.79 \\
ER+SL & 76.96 & 69.46 & 73.93 & 67.32 & 62.59 & 70.89 & 77.86 & 72.05 & 83.93 & 83.39 & 73.84 & 14.81 \\
ER+JoCoR & 84.64 & 78.66 & 79.64 & 78.93 & 71.96 & 79.20 & 86.61 & \textbf{82.41} & 84.02 & 91.34 & 81.74 & 11.54 \\
I-LoRA+SL & 88.30 & 83.04 & 80.89 & 81.34 & 73.75 & 80.27 & 83.30 & 75.62 & 85.71 & 86.96 & 81.92 & 9.38 \\
I-LoRA+JoCoR & 73.21 & 66.32 & 64.56 & 70.12 & 60.32 & 68.57 & 72.6 & 55.74 & 58.78 & 67.91 & 65.81 & 11.82 \\
\hline
\texttt{RiCL } & 83.66 & \textbf{82.14} & \textbf{85.00} & \textbf{83.84} & \textbf{79.73} & \textbf{85.36} & \textbf{88.57} & 77.68 & \textbf{85.89} & \textbf{95.80} & \textbf{84.77} & \textbf{7.9} \\ \hline
Multitask & 95.62 & 90.62 & 89.38 & 92.23 & 85.45 & 86.96 & 91.7 & 85.54 & 92.59 & 92.50 & 90.26 & - \\
Multitask+SL & 96.34 & 92.05 & 90.62 & 93.3 & 85.98 & 86.79 & 91.88 & 86.61 & 92.95 & 92.95 & 90.95 & - \\
\specialrule{1.5pt}{0pt}{0pt}
\end{tabular}}
\caption{Performance (\%) of our \texttt{RiCL} and distinct continual learning method on Fewrel. The best results are emphasized in \textbf{bold}. Multitask (+SL) is the upper bound.}
\label{tab_Fewrel}
\vspace{-1mm}
\end{table*}

\subsection{Main Results}

In this section, we compare \texttt{RiCL} with strong continual‑learning and noisy‑label baselines on Tacred and Fewrel (Tables \ref{tab_Tacred}–\ref{tab_Fewrel}) under 20\% symmetric label noise. Three findings stand out.
\textbf{First}, \texttt{RiCL} outperforms all the strong baselines in terms of AP and AF over both Tacred and Fewrel. \texttt{RiCL} achieves 82.70\% AP with just 1.69\% AF on Tacred dataset, clearly beating the best hybrid baseline (I-LoRA + SL) at 74.31\% AP and 5.88\% AF, offering 8.4\% more AP while reducing forgetting by four times. In contrast, S6, an online noisy-label method, reaches only 70.51\% AP and does not report AF because it uses offline fine-tuning after the final task. A similar pattern appears on Fewrel, where \texttt{RiCL} maintains its lead with 84.77\% AP and 7.90\% AF, surpassing I-LoRA + SL (81.92\% AP, 9.38\% AF) and S6 (72.21\% AP).
\textbf{Second}, simply adding noise-label learning methods to continual learning approaches does little to solve the issues of label noise and forgetting. Hybrid methods like ER + SL, ER + JoCoR, or I-LoRA + SL provide only slight accuracy improvements but still suffer from significant forgetting. \texttt{RiCL} performs better than the best noisy label learning-enhanced methods over Tacred and Fewrel (74.31\% vs. 82.70\% and 81.92\% vs. 84.77\% in terms of AP). 
\textbf{Third}, these results show that \texttt{RiCL}'s joint, feedback-focused design is essential for tackling forgetting and noise issues in the interactive continual learning setting. 
It narrows the gap to the upper bound baseline multitask to a few points, but there remains room for improvement, especially in further reducing forgetting and enhancing robustness against more complex noise patterns.


\subsection{Ablation Studies}
\begin{wraptable}{r}{0.45\textwidth}
    \centering
    \vspace{-3mm}
    \setlength{\tabcolsep}{0.8mm}{
    \begin{tabular}{ccccccc}
        \specialrule{1.5pt}{0pt}{0pt}
        \multirow{2}{*}{\centering TCP} & \multirow{2}{*}{\centering NCL} & \multirow{2}{*}{\centering IPO} & \multicolumn{2}{c}{Tacred} & \multicolumn{2}{c}{Fewrel} \\
        \cmidrule(lr){4-5} \cmidrule(lr){6-7}
                &                      &     & AP & AF & AP & AF \\
        \hline
        \ding{51} & \ding{51}             & \ding{51} & \textbf{82.70} & \textbf{1.69} & \textbf{84.77} & \textbf{7.9} \\
        \ding{51} & \ding{51}              & \ding{55} & 82.53 & 6.32 & 84.12 & 10.1 \\
        \ding{51} & \ding{55}              & \ding{51} & 82.41 & 4.51 & 84.22 & 9.2 \\
        \ding{55} &  \ding{51}            & \ding{51} & 81.16 & 6.82 & 83.05 & 9.61 \\
        \specialrule{1.5pt}{0pt}{0pt}
    \end{tabular}}
    \caption{The results of ablation studies.}
    \label{tab:ablation_study}
    \vspace{-3mm}
\end{wraptable}
To assess the contribution of each component in our proposed framework, we conduct an ablation study as shown in Table~\ref{tab:ablation_study}. We focus on three key modules: the Temporal Consistency-aware Purifier (TCP), Noise-resistant Contrastive Learning (NCL), and Interaction-aware Preference Optimization (IPO). 
TCP is the indispensable backbone: once it is disabled, both accuracy and forgetting deteriorate sharply because all subsequent learning must rely on corrupted signals. 
With TCP in place, the other two components target specific weaknesses. Without IPO, accuracy stays nearly the same, but forgetting jumps significantly, showing that aligning the model with user preferences is key to retaining what it has learned. On the other hand, removing NCL slightly reduces accuracy and moderately increases forgetting, suggesting that contrastive learning mainly helps the model stay stable despite noisy labels. In short, TCP cleans incoming data, NCL strengthens reliable features, and IPO ensures the model stays in line with user expectations—together, they achieve the best balance between accuracy and memory retention.


\begin{wraptable}{r}{0.67\textwidth}
\centering
\small
\vspace{-4mm}
\setlength{\tabcolsep}{0.7mm}{
\begin{tabular}{lcccccccc}
\specialrule{1.5pt}{0pt}{0pt}
\multirow{2}{*}{Methods} & \multicolumn{4}{c}{\textbf{Tacred}} & \multicolumn{4}{c}{\textbf{Fewrel}} \\
\cmidrule(lr){2-5}\cmidrule(lr){6-9}
 & \multicolumn{2}{c}{20\% Noise} & \multicolumn{2}{c}{40\% Noise} & \multicolumn{2}{c}{20\% Noise} & \multicolumn{2}{c}{40\% Noise} \\
\cmidrule(lr){2-3}\cmidrule(lr){4-5}\cmidrule(lr){6-7}\cmidrule(lr){8-9}
 & AP $\uparrow$ & AF $\downarrow$ & AP $\uparrow$ & AF $\downarrow$ & AP $\uparrow$ & AF $\downarrow$ & AP $\uparrow$ & AF $\downarrow$ \\
\hline
SeqFT           & 64.69 & 30.44  & 59.66 & 29.29  & 68.09 & 28.19  & 57.11 & 28.41  \\
SeqFT + SL      & 66.11 & 31.62  & 60.70 & 31.38  & 70.14 & 26.43  & 59.76 & 29.30  \\
ER + JoCoR         & 72.98 & 15.00 & 58.43 & 19.87 & 81.74 & 11.54 & 63.08 & 18.28 \\
MIR + JoCoR        & 70.04 & 19.61 & 59.73 & 23.52 & 78.89 & 14.79 & 64.45 & 18.89 \\
I-LoRA + JoCoR      & 71.39 & 7.68  & 52.61 & 14.15  & 76.95 & 8.32  & 65.81 & 11.82  \\
\hline
\texttt{RiCL}      & 82.70 & 1.69  & 79.58 & -3.81 & 84.77 & 7.90  & 83.78 & 8.86  \\ \hline
Multitask          & 83.63 & -  & 77.19 & -  & 90.26 & -  & 82.55 & -  \\
Multitask + SL     & 84.02 & -  & 80.17 & -  & 90.95 & -  & 86.09 & -  \\
\specialrule{1.5pt}{0pt}{0pt}
\end{tabular}}
\caption{Average performance (AP) and average forgetting (AF) on Tacred and Fewrel under 20\% and 40\% symmetric label noise. 
}
\label{tab:different_noise_rate}
\vspace{-3mm}
\end{wraptable}

\subsection{Further Analysis}
\paragraph{Influence of Noise Ratio}
To quantify how label noise affects continual‑learning performance, we inject two noise levels (20\% and 40\%) into Tacred and Fewrel and report AP and AF in Table \ref{tab:different_noise_rate}. Detailed per-task accuracies under 40\% noise are presented in Table \ref{tab_40_Tacred} for Tacred and Table \ref{tab_40_Fewrel} for Fewrel in the Appendix.
From the results, we find that our model performs better than all the baselines in terms of AP and AF.
Moreover, \texttt{RiCL} stays close to multitask performance and even gains slightly at higher noise levels, thanks to its temporal purifier, preference optimization, and contrastive learning. 
For example, the best AP with 40\% noise is 65.81\% on Fewrel while our model is 83.78\%, which is close to the upper bound 86.09\%.
Though increasing the noise reduces the performance, the drops of our model are limited (84.77\% vs 83.78\% in terms of AP and 7.90\% vs 8.86\% in terms of AF).
Additionally, we observe that JoCoR-enhanced baselines sometimes suffer sharp accuracy drops and increased forgetting, even with the noisy learning strategy. 
These results show that real-time noise handling is essential for effective continual learning in noisy, streaming environments.

\begin{wraptable}{r}{0.40\textwidth}
\vspace{-4mm}
\centering
\small
\setlength{\tabcolsep}{2mm}{
\begin{tabular}{ccc}
\specialrule{1.5pt}{0pt}{0pt}
Noise Rate (\%) & $AP\,(\uparrow)$ & $AF\,(\downarrow)$ \\ \hline
20 & 82.70 & 1.69  \\
30 & 81.21 & 4.18  \\
40 & 78.69 & 3.7 \\
50 & 73.95 & -2.18 \\ 
\specialrule{1.5pt}{0pt}{0pt}
\end{tabular}}
\vspace{-1mm}
\caption{Performance (\%) of \texttt{RiCL} under different noise rates on Tacred.}
\label{table:different_noise_Tacred}
  \vspace{-2mm}
\end{wraptable}
Furthermore, we explore the performance of \texttt{RiCL} with the label-noise rate from 20\% to 50\% (Table \ref{table:different_noise_Tacred}). Detailed per-task accuracies under each noise level are provided in Appendix Table \ref{talbe_different_noise_Tacred_1}.
It is intuitive that the model's performance deteriorates as noise increases.
However, we find that the performance drops only a little from 20\% to 30\% noise. 
Even when 50\% noise (where half the labels are wrong), it still retains around 90\% of its 20\%-noise performance.
AF rises from 1.69 at 20\% noise to 4.18 at 30\%, reflecting increased forgetting. However, it unexpectedly declines at higher noise levels, even turning negative (-2.18) at 50\%.
This decline occurs because immediate post-task accuracy ($M_i^i$) has already degraded under severe noise. As noise rises, AF therefore becomes unreliable, whereas AP consistently reflects \texttt{RiCL}’s stability.

\begin{wrapfigure}{r}{0.60\textwidth}
  \vspace{-3mm}  
  \centering
  \includegraphics[width=0.58\textwidth]{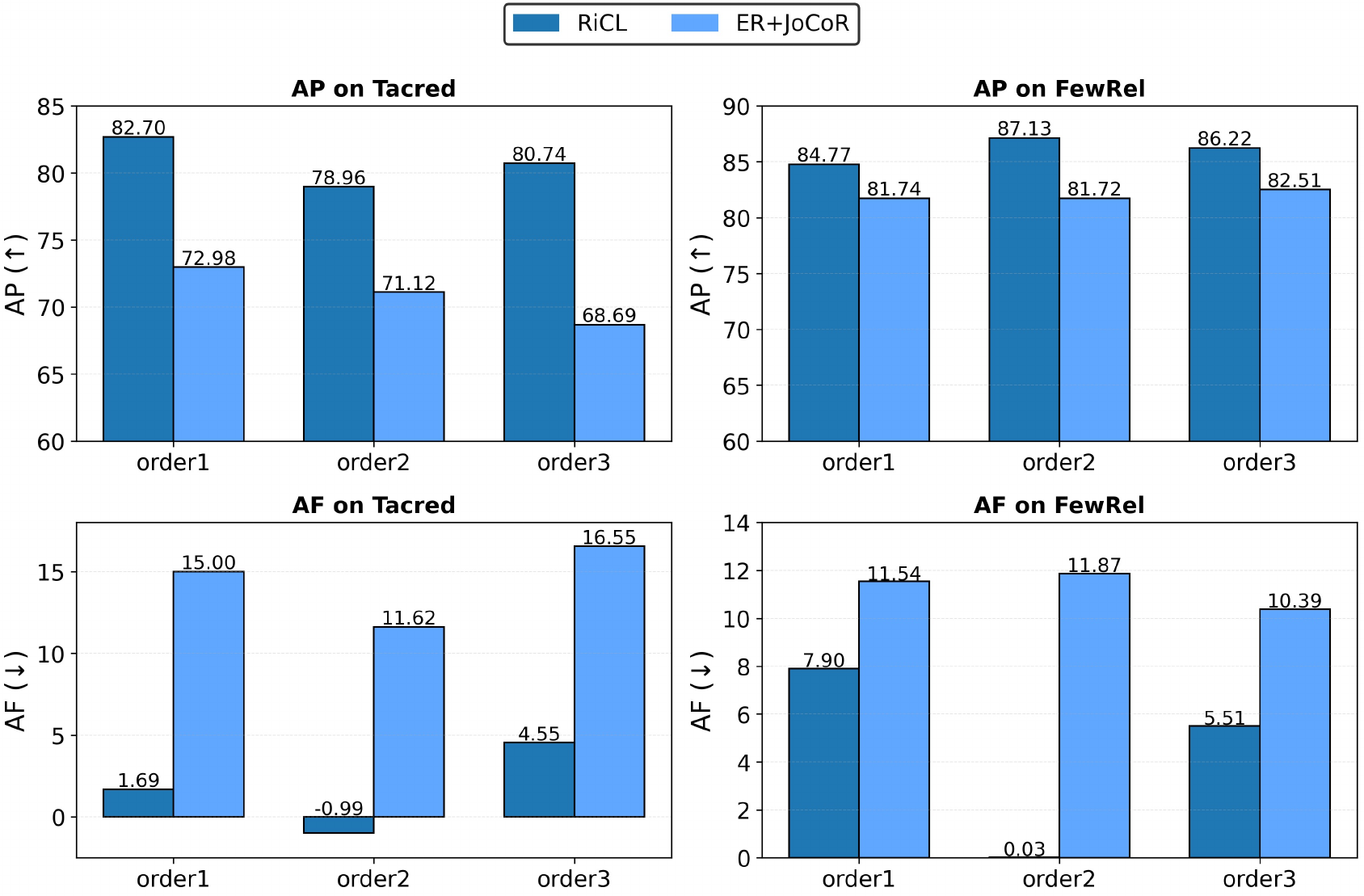}
  \vspace{-1mm}
  \caption{AP and AF on Tacred and Fewrel across distinct task orders.}
  \label{fig:task_order}
  \vspace{-2mm}
\end{wrapfigure}

\paragraph{Influence of Task Order}
Figure \ref{fig:task_order} shows how task order affects each method. We try three streams: the original order (0→1→…→9), the exact reverse (9→8→…→0), and a purposely mixed shuffle (3 → 7 → 2 → 8 → 5 → 1 → 9 → 0 → 6 → 4).  Detailed per-task accuracies for each order are presented in Appendix Table \ref{tab_Tacred_order} (Tacred) and Table \ref{tab_Fewrel_order} (Fewrel).
We find that our \texttt{RiCL} model outperforms the baseline over all the orders. For example, \texttt{RiCL} obtains more than seven points on order2 in terms of AP.
Additionally, the performance of the baseline is is sensitive to task order while our model performs stably across different orders 
\texttt{RiCL}’s accuracy stays almost unchanged and forgetting stays low no matter which order we use. By contrast, the accuracy of the replay baseline ER + JoCoR drops and forgetting rises when the tasks are reversed or shuffled. This shows \texttt{RiCL} can handle whatever task sequence comes its way—an important trait for real-world systems where new tasks rarely arrive in a tidy timeline.



\noindent

%% file: section/appendix.tex
\section*{Appendix}
\renewcommand{\thesubsection}{A.\arabic{subsection}}

\subsection{Implementation Details}
\label{app:impl}
To optimize the training process effectively, we employ distinct learning rates for different components. Specifically, the learning rate for TCP (\(\mathcal{M}_p\)) is set to \(1 \times 10^{-4}\). For the noise-resistant contrastive learning of the primary LLM \(\mathcal{M}\), the learning rate is set to \(1 \times 10^{-5}\), while the interaction-aware direct preference optimization of \(\mathcal{M}\) adopts a learning rate of \(5 \times 10^{-6}\). Additionally, the number of alternative labels is fixed at \(L = 5\).

\vspace{3pt}
\subsection{Noise-Level Sensitivity on Tacred}
\label{app:noise_Tacred}
Table \ref{talbe_different_noise_Tacred_1} reports the per-task classification accuracy of \texttt{RiCL} on Tacred under four symmetric-noise settings (20\%–50\%). The ten central columns correspond to Tasks 1–10, while the final two columns list the overall mean AP and AF for each noise level.
\begin{table*}[htbp]
\centering
\small
\setlength{\tabcolsep}{0.6mm}{
\begin{tabular}{ccccccccccccc}
\specialrule{1.5pt}{0pt}{0pt}
\multirow{2}{*}{Noise Rate(\%)} & \multicolumn{10}{c}{Various task in Tacred}                                                                                                                             & \multirow{2}{*}{$AP(\uparrow)$} & \multicolumn{1}{l}{\multirow{2}{*}{$AF(\downarrow)$}} \\ \cline{2-11}
                         & Task 1           & Task 2           & Task 3           & Task 4           & Task 5           & Task 6           & Task 7           & Task 8           & Task 9           & Task 10           &                     & \multicolumn{1}{l}{}                     \\ \hline
20 & 85.09 & 90.21 & 84.67 & 80.37 & 87.94 & 78.84 & 77.41 & 78.92 & 78.02 & 85.51 & 82.70 & 1.69 \\
30 & 81.37 & 88.11 & 81.02 & 77.91 & 90.07 & 75.34 & 67.39 & 75.0 & 84.62 & 91.3 & 81.21 & 4.18 \\
40 & 84.47 & 89.51 & 81.02 & 78.53 & 86.52 & 73.29 & 61.96 & 72.41 & 78.02 & 81.16 & 78.69 & 3.7 \\
50 & 61.49 & 83.92 & 82.48 & 66.26 & 86.52 & 68.49 & 56.52 & 73.28 & 73.63 & 86.96 & 73.95 & -2.18 \\

\specialrule{1.5pt}{0pt}{0pt}
\end{tabular}}
\caption{Performance (\%) of our \texttt{RiCL} and distinct noise rate on Tacred. We list the accuracy for each task along with $AP$ and $AF$.}
\label{talbe_different_noise_Tacred_1}
\end{table*}

\subsection{Per-Task Performance on Tacred (40\% Noise)}
\label{app:perf_Tacred40}
Table \ref{tab_40_Tacred} lists the complete per-task classification accuracies for all evaluated continual-learning methods on the Tacred dataset under a 40\% symmetric label-noise setting. The ten central columns correspond to Tasks 1–10, while the two rightmost columns provide the overall mean AP and AF. 
\begin{table*}[htbp]
\centering
\small
\setlength{\tabcolsep}{0.6mm}{
\begin{tabular}{lcccccccccccc}
\specialrule{1.5pt}{0pt}{0pt}
\multirow{2}{*}{Methods} & \multicolumn{10}{c}{Various task in Tacred(40\% noise)}                                                                                                                             & \multirow{2}{*}{$AP(\uparrow)$} & \multicolumn{1}{l}{\multirow{2}{*}{$AF(\downarrow)$}} \\ \cline{2-11}
                         & Task 1           & Task 2           & Task 3           & Task 4           & Task 5           & Task 6           & Task 7           & Task 8           & Task 9           & Task 10           &                     & \multicolumn{1}{l}{}                     \\ \hline
Multitask & 85.09 & 80.42 & 78.83 & 75.46 & 85.82 & 73.29 & 70.65 & 68.97 & 78.02 & 75.36 & 77.19 & - \\
Multitask+SL & 84.47 & 88.81 & 78.83 & 77.3 & 90.07 & 77.4 & 73.91 & 69.83 & 81.32 & 79.71 & 80.17 & - \\
SeqFT & 38.51 & 60.14 & 67.88 & 48.47 & 85.82 & 60.27 & 29.35 & 63.79 & 62.64 & 79.71 & 59.66 & 29.29 \\
SeqFT+SL & 39.13 & 57.34 & 69.34 & 44.79 & 81.56 & 61.64 & 20.65 & 72.41 & 70.33 & 89.86 & 60.70 & 31.38 \\
 \hline
ER+JoCoR & 49.69 & 65.03 & 68.61 & 33.74 & 63.83 & 47.95 & 52.17 & 57.76 & 57.14 & 88.41 & 58.43 & 19.87 \\
MIR+JoCoR & 47.83 & 52.45 & 66.42 & 49.08 & 64.54 & 43.84 & 60.87 & 53.45 & 74.73 & 84.06 & 59.73 & 23.52 \\
I-LORA+JoCoR & 49.69 & 57.34 & 58.39 & 44.17 & 58.16 & 35.62 & 54.35 & 34.48 & 52.75 & 81.16 & 52.61 & 14.15 \\
\hline
\texttt{RiCL}  & 84.47 & 89.51 & 81.02 & 78.53 & 86.52 & 73.29 & 61.96 & 72.41 & 78.02 & 81.16 & 78.69 & 3.7 \\
\specialrule{1.5pt}{0pt}{0pt}
\end{tabular}}
\caption{Performance (\%) of our \texttt{RiCL} and distinct continual learning method on Tacred. We list the accuracy for each task along with $AP$ and $AF$.}
\label{tab_40_Tacred}
\end{table*}
\subsection{Per-Task Performance on Fewrel (40\% Noise)}
\label{app:perf_Fewrel40}
Table \ref{tab_40_Fewrel} lists the complete per-task classification accuracies for all evaluated continual-learning methods on the Fewrel dataset under a 40\% symmetric label-noise setting. The ten central columns correspond to Tasks 1–10, while the two rightmost columns provide the overall mean AP and AF. 
\begin{table*}[htbp]
\centering
\small
\setlength{\tabcolsep}{0.6mm}{
\begin{tabular}{lcccccccccccc}
\specialrule{1.5pt}{0pt}{0pt}
\multirow{2}{*}{Methods} & \multicolumn{10}{c}{Various task in Fewrel(40\% noise)}                                                                                                                             & \multirow{2}{*}{$AP(\uparrow)$} & \multicolumn{1}{l}{\multirow{2}{*}{$AF(\downarrow)$}} \\ \cline{2-11}
                         & Task 1           & Task 2           & Task 3           & Task 4           & Task 5           & Task 6           & Task 7           & Task 8           & Task 9           & Task 10           &                     & \multicolumn{1}{l}{}                     \\ \hline
Multitask & 88.21 & 83.3 & 80.09 & 84.02 & 77.32 & 78.3 & 85.98 & 77.77 & 85.8 & 84.73 & 82.55 & - \\
Multitask+SL & 92.32 & 87.68 & 85.71 & 86.88 & 78.93 & 81.88 & 88.3 & 80.89 & 89.2 & 89.11 & 86.09 & - \\
SeqFT & 55.71 & 47.86 & 57.59 & 48.3 & 46.52 & 54.29 & 61.07 & 50.00 & 70.00 & 79.73 & 57.11 & 28.41 \\
SeqFT+SL & 56.16 & 51.25 & 59.91 & 52.86 & 51.52 & 58.21 & 63.84 & 47.23 & 73.39 & 83.21 & 59.76 & 29.30 \\
\hline

ER+JoCoR & 67.86 & 58.04 & 66.43 & 51.61 & 56.07 & 74.2 & 68.66 & 50.0 & 58.57 & 79.38 & 63.08 & 18.28 \\

MIR+JoCoR & 60.18 & 62.86 & 67.05 & 54.2 & 59.82 & 68.39 & 67.05 & 58.3 & 63.66 & 83.04 & 64.45 & 18.89 \\
I-LORA+JoCoR & 37.5 & 33.48 & 55.89 & 41.43 & 38.3 & 63.48 & 60.62 & 33.66 & 35.98 & 48.39 & 44.87 & 11.98 \\  
\hline
\texttt{RiCL}  & 91.25 & 81.16 & 81.96 & 83.12 & 74.55 & 83.12 & 89.55 & 72.86 & 87.5 & 92.77 & 83.78 & 8.86 \\
\specialrule{1.5pt}{0pt}{0pt}
\end{tabular}}
\caption{Performance (\%) of our \texttt{RiCL} and distinct continual learning method on Fewrel. We list the accuracy for each task along with $AP$ and $AF$.}
\label{tab_40_Fewrel}
\end{table*}

\subsection{Robustness to Curriculum Order}
\label{app:order_robust}
Tables \ref{tab_Tacred_order} and \ref{tab_Fewrel_order} report the per-task accuracies of \texttt{RiCL} and the ER + JoCoR baseline under three distinct task orders on Tacred and Fewrel, respectively; each table lists post-training accuracy for Tasks 1–10 alongside the overall mean AP and AF for every method–order pair.

\begin{table*}[htbp]
\centering
\small
\setlength{\tabcolsep}{0.5mm}{
\begin{tabular}{cccccccccccccc}
\specialrule{1.5pt}{0pt}{0pt}
\multirow{2}{*}{Task Order} & \multirow{2}{*}{Methods} & \multicolumn{10}{c}{Various task in Tacred}                                                                                                                             & \multirow{2}{*}{$AP(\uparrow)$} & \multicolumn{1}{l}{\multirow{2}{*}{$AF(\downarrow)$}} \\ \cline{3-12}
         &                & Task 1           & Task 2           & Task 3           & Task 4           & Task 5           & Task 6           & Task 7           & Task 8           & Task 9           & Task 10           &                     & \multicolumn{1}{l}{}                     \\ \hline
\multirow{2}{*}{Order 1} & \texttt{ER+JoCoR} & 75.78 & 76.92 & 75.91 & 55.83 & 80.85 & 71.92 & 70.65 & 64.66 & 70.33 & 86.96 & 72.98 & 15.00 \\
&  \texttt{RiCL} & 85.09 & 90.21 & 84.67 & 80.37 & 87.94 & 78.84 & 77.41 & 78.92 & 78.02 & 85.51 & 82.70 & 1.69 \\ \hline
\multirow{2}{*}{Order 2} & \texttt{ER+JoCoR}& 78.82 & 73.22 & 72.48 & 73.01 & 78.65 & 59.59 & 63.04 & 62.07 & 73.52 & 76.81 & 71.12 & 11.62 \\
&  \texttt{RiCL} & 89.44 & 94.41 & 89.05 & 81.6 & 91.49 & 71.92 & 73.91 & 69.83 & 71.43 & 56.52 & 78.96 & -0.99 \\ \hline
\multirow{2}{*}{Order 3} & \texttt{ER+JoCoR} & 82.42 & 67.70 & 60.58 & 61.96 & 76.92 & 56.90 & 85.11 & 52.76 & 76.81 & 65.75 & 68.69 & 16.55\\
& \texttt{RiCL} & 76.92 & 91.93 & 82.48 & 75.00 & 90.91 & 65.52 & 92.91 & 71.17 & 81.16 & 79.45 & 80.74 & 4.55  \\
\specialrule{1.5pt}{0pt}{0pt}
\end{tabular}}
\caption{Performance (\%) of \texttt{RiCL} and ER+JoCoR across distinct task orders on Tacred. We list the accuracy for each task along with $AP$ and $AF$.}
\label{tab_Tacred_order}
\end{table*}

\begin{table*}[htbp]
\centering
\small
\setlength{\tabcolsep}{0.5mm}{
\begin{tabular}{cccccccccccccc}
\specialrule{1.5pt}{0pt}{0pt}
\multirow{2}{*}{Task Order} & \multirow{2}{*}{Task Order} & \multicolumn{10}{c}{Various task in Fewrel}                                                                                                                             & \multirow{2}{*}{$AP(\uparrow)$} & \multicolumn{1}{l}{\multirow{2}{*}{$AF(\downarrow)$}} \\ \cline{3-12}
       &                  & Task 1           & Task 2           & Task 3           & Task 4           & Task 5           & Task 6           & Task 7           & Task 8           & Task 9           & Task 10           &                     & \multicolumn{1}{l}{}                     \\ \hline
\multirow{2}{*}{Order 1}  & \texttt{ER+JoCoR} & 84.64 & 78.66 & 79.64 & 78.93 & 71.96 & 79.20 & 86.61 & 82.41 & 84.02 & 91.34 & 81.74 & 11.54 \\
 & \texttt{RiCL}& 83.66 & 82.14 & 85.00 & 83.84 & 79.73 & 85.36 & 88.57 & 77.68 & 85.89 & 95.80 & 84.77 & 7.9 \\ \hline
\multirow{2}{*}{Order 2}  & \texttt{ER+JoCoR} & 96.07 & 82.68 & 85.98 & 88.04 & 73.21 & 79.11 & 79.20 & 74.46 & 83.21 & 75.27 & 81.72 & 11.87 \\
 & \texttt{RiCL}& 98.12 & 89.55 & 90.45 & 87.5 & 80.62 & 85.27 & 83.93 & 79.38 & 88.12 & 88.39 & 87.13 & 0.03\\ \hline
\multirow{2}{*}{Order 3} & \texttt{ER+JoCoR} & 84.11 & 90.54 & 81.70 & 85.0 & 78.84 & 73.48 & 85.8 & 79.46 & 80.62 & 85.54 & 82.51 & 10.39  \\
 & \texttt{RiCL} & 86.52 & 87.32 & 82.14 & 89.64 & 76.61 & 84.11 & 94.46 & 90.71 & 81.7 & 89.02 & 86.22 & 5.51  \\
\specialrule{1.5pt}{0pt}{0pt}
\end{tabular}}
\caption{Performance (\%) of our \texttt{RiCL} and ER+JoCoR across distinct task orders on Fewrel. We list the accuracy for each task along with $AP$ and $AF$.}
\label{tab_Fewrel_order}
\end{table*}



\begin{table*}[t]
  \centering
  \small
  \setlength{\tabcolsep}{1.5mm}{
  \begin{tabular}{lccccccc}
   \specialrule{1.5pt}{0pt}{0pt}
    \multirow{2}{*}{Dataset} & \multirow{2}{*}{Order} 
        & \multicolumn{2}{c}{\textbf{RiCL}} 
        & \multicolumn{2}{c}{\textbf{ER{+}JoCoR}} 
        & \multirow{2}{*}{$\Delta$AP\,$\uparrow$} 
        & \multirow{2}{*}{$\Delta$AF\,$\downarrow$} \\ 
    \cmidrule(lr){3-4}\cmidrule(lr){5-6}
        & & AP\,$\uparrow$ & AF\,$\downarrow$ & AP\,$\uparrow$ & AF\,$\downarrow$ & \\ 
    \hline
    \multirow{3}{*}{Tacred} 
        & order\,1 & 82.70 &  1.69 & 72.98 & 15.00 & +9.72 & -13.31 \\ 
        & order\,2 & 78.96 & –0.99 & 71.12 & 11.62 & +7.84 & -12.61 \\ 
        & order\,3 & 80.74 &  4.55 & 68.69 & 16.55 & +12.05 & -12.00 \\[2pt]
    \hline
    \multirow{3}{*}{Fewrel} 
        & order\,1 & 84.77 &  7.90 & 81.74 & 11.54 & +3.03 & -3.64 \\ 
        & order\,2 & 87.13 &  0.03 & 81.72 & 11.87 & +5.41 & -11.54 \\ 
        & order\,3 & 86.22 &  5.51 & 82.51 & 10.39 & +3.71 & -4.88 \\ 
   \specialrule{1.5pt}{0pt}{0pt}
  \end{tabular}}
   \caption{Influence of task order on \texttt{RiCL} compared with the noisy‑label baseline \texttt{ER+JoCoR}. 
   }
  \label{tab:task_order}
\end{table*}

\begin{figure*}[t]
  \centering
  \includegraphics[width=0.9\textwidth]{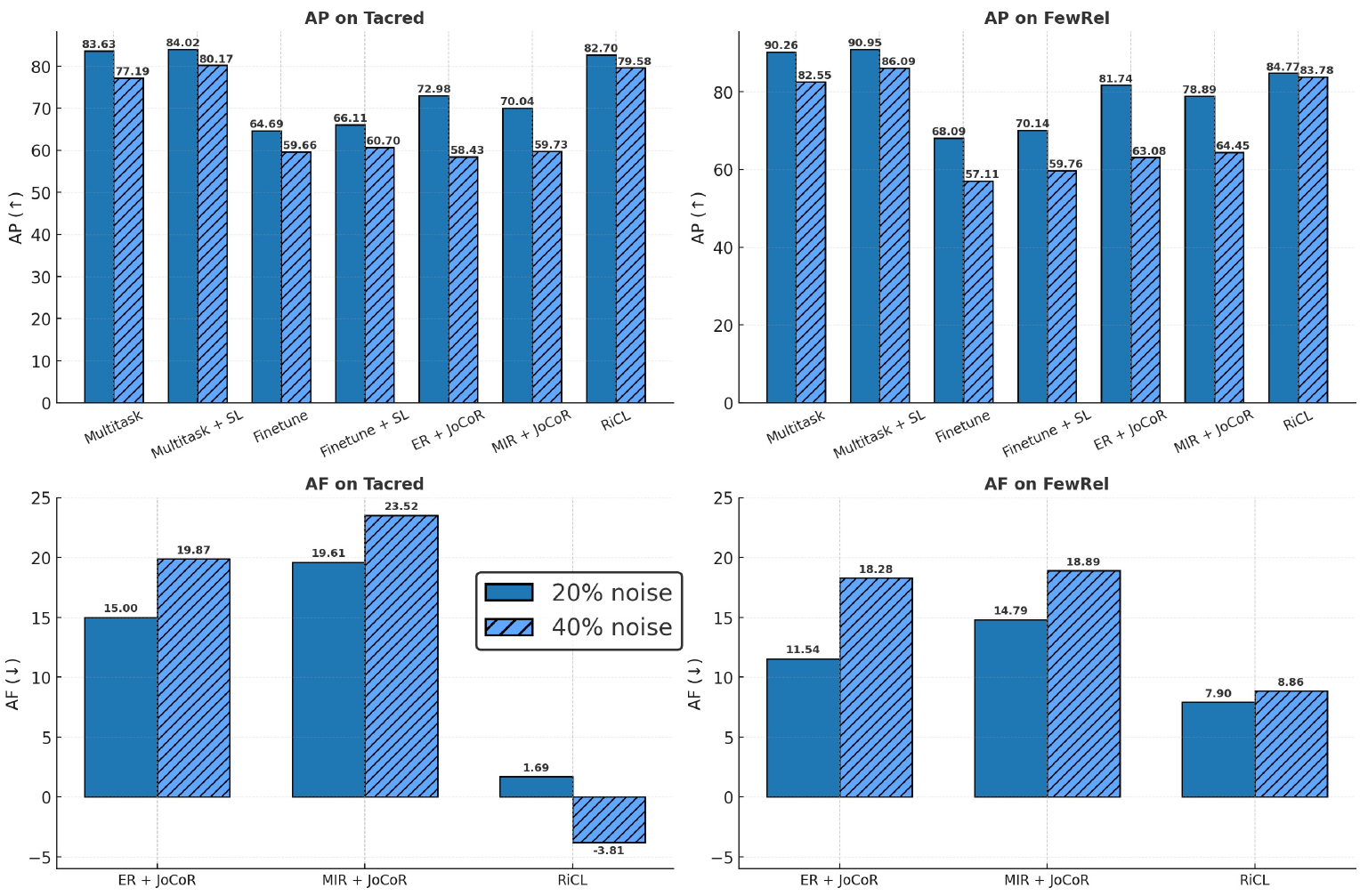}
  \caption{AP and AF on Tacred and Fewrel under 20\% and 40\% symmetric label noise.}
  \label{fig:different_noise_rate}
\end{figure*}

%% file: arxiv.bbl
\begin{thebibliography}{10}

\bibitem{wang2024comprehensive}
Liyuan Wang, Xingxing Zhang, Hang Su, and Jun Zhu.
\newblock A comprehensive survey of continual learning: Theory, method and application.
\newblock {\em IEEE Transactions on Pattern Analysis and Machine Intelligence}, 2024.

\bibitem{li2017learning}
Zhizhong Li and Derek Hoiem.
\newblock Learning without forgetting.
\newblock {\em TPAMI}, 40(12):2935--2947, 2017.

\bibitem{kirkpatrick2017overcoming}
James Kirkpatrick, Razvan Pascanu, Neil Rabinowitz, Joel Veness, Guillaume Desjardins, Andrei~A Rusu, Kieran Milan, John Quan, Tiago Ramalho, Agnieszka Grabska-Barwinska, et~al.
\newblock Overcoming catastrophic forgetting in neural networks.
\newblock {\em NAS}, 114(13):3521--3526, 2017.

\bibitem{de2021continual}
Matthias De~Lange, Rahaf Aljundi, Marc Masana, Sarah Parisot, Xu~Jia, Ale{\v{s}} Leonardis, Gregory Slabaugh, and Tinne Tuytelaars.
\newblock A continual learning survey: Defying forgetting in classification tasks.
\newblock {\em IEEE transactions on pattern analysis and machine intelligence}, 44(7):3366--3385, 2021.

\bibitem{shaheen2022continual}
Khadija Shaheen, Muhammad~Abdullah Hanif, Osman Hasan, and Muhammad Shafique.
\newblock Continual learning for real-world autonomous systems: Algorithms, challenges and frameworks.
\newblock {\em Journal of Intelligent \& Robotic Systems}, 105(1):9, 2022.

\bibitem{liu2021lifelong}
Bing Liu and Sahisnu Mazumder.
\newblock Lifelong and continual learning dialogue systems: learning during conversation.
\newblock In {\em AAAI}, volume~35, pages 15058--15063, 2021.

\bibitem{yang2025recent}
Yutao Yang, Jie Zhou, Xuanwen Ding, Tianyu Huai, Shunyu Liu, Qin Chen, Yuan Xie, and Liang He.
\newblock Recent advances of foundation language models-based continual learning: A survey.
\newblock {\em ACM Computing Surveys}, 57(5):1--38, 2025.

\bibitem{huai2025cl}
Tianyu Huai, Jie Zhou, Xingjiao Wu, Qin Chen, Qingchun Bai, Ze~Zhou, and Liang He.
\newblock Cl-moe: Enhancing multimodal large language model with dual momentum mixture-of-experts for continual visual question answering.
\newblock {\em arXiv preprint arXiv:2503.00413}, 2025.

\bibitem{rolnick2019experience}
David Rolnick, Arun Ahuja, Jonathan Schwarz, Timothy Lillicrap, and Gregory Wayne.
\newblock Experience replay for continual learning.
\newblock {\em Advances in neural information processing systems}, 32, 2019.

\bibitem{ahn2019uncertainty}
Hongjoon Ahn, Sungmin Cha, Donggyu Lee, and Taesup Moon.
\newblock Uncertainty-based continual learning with adaptive regularization.
\newblock {\em Advances in neural information processing systems}, 32, 2019.

\bibitem{ding2024boosting}
Xuanwen Ding, Jie Zhou, Liang Dou, Qin Chen, Yuanbin Wu, Arlene Chen, and Liang He.
\newblock Boosting large language models with continual learning for aspect-based sentiment analysis.
\newblock In {\em Findings of the Association for Computational Linguistics: EMNLP 2024}, pages 4367--4377, 2024.

\bibitem{bidaki2025onlinecontinuallearningsystematic}
Seyed~Amir Bidaki, Amir Mohammadkhah, Kiyan Rezaee, Faeze Hassani, Sadegh Eskandari, Maziar Salahi, and Mohammad~M. Ghassemi.
\newblock Online continual learning: A systematic literature review of approaches, challenges, and benchmarks, 2025.

\bibitem{mai2022online}
Zheda Mai, Ruiwen Li, Jihwan Jeong, David Quispe, Hyunwoo Kim, and Scott Sanner.
\newblock Online continual learning in image classification: An empirical survey.
\newblock {\em Neurocomputing}, 469:28--51, 2022.

\bibitem{wang2023cba}
Quanziang Wang, Renzhen Wang, Yichen Wu, Xixi Jia, and Deyu Meng.
\newblock Cba: Improving online continual learning via continual bias adaptor.
\newblock In {\em ICCV}, pages 19082--19092, 2023.

\bibitem{qi2024interactive}
Biqing Qi, Xinquan Chen, Junqi Gao, Dong Li, Jianxing Liu, Ligang Wu, and Bowen Zhou.
\newblock Interactive continual learning: Fast and slow thinking.
\newblock In {\em Proceedings of the IEEE/CVF Conference on Computer Vision and Pattern Recognition}, pages 12882--12892, 2024.

\bibitem{wang2023orthogonal}
Xiao Wang, Tianze Chen, Qiming Ge, Han Xia, Rong Bao, Rui Zheng, Qi~Zhang, Tao Gui, and Xuan-Jing Huang.
\newblock Orthogonal subspace learning for language model continual learning.
\newblock In {\em Findings of the Association for Computational Linguistics: EMNLP 2023}, pages 10658--10671, 2023.

\bibitem{shin2017continual}
Hanul Shin, Jung~Kwon Lee, Jaehong Kim, and Jiwon Kim.
\newblock Continual learning with deep generative replay.
\newblock {\em Advances in neural information processing systems}, 30, 2017.

\bibitem{van2020brain}
Gido~M Van~de Ven, Hava~T Siegelmann, and Andreas~S Tolias.
\newblock Brain-inspired replay for continual learning with artificial neural networks.
\newblock {\em Nature communications}, 11(1):4069, 2020.

\bibitem{du2023static}
Mingzhe Du, Anh~Tuan Luu, Bin Ji, and See-kiong Ng.
\newblock From static to dynamic: A continual learning framework for large language models.
\newblock {\em arXiv preprint arXiv:2310.14248}, 2023.

\bibitem{sun2019lamol}
Fan{-}Keng Sun, Cheng{-}Hao Ho, and Hung{-}Yi Lee.
\newblock {LAMOL:} language modeling for lifelong language learning.
\newblock In {\em ICLR}. OpenReview.net, 2020.

\bibitem{wang2022rvae}
Han Wang, Ruiliu Fu, Xuejun Zhang, and Jun Zhou.
\newblock Rvae-lamol: Residual variational autoencoder to enhance lifelong language learning.
\newblock In {\em IJCNN}, pages 1--9. IEEE, 2022.

\bibitem{huang2021continual}
Yufan Huang, Yanzhe Zhang, Jiaao Chen, Xuezhi Wang, and Diyi Yang.
\newblock Continual learning for text classification with information disentanglement based regularization.
\newblock In {\em Proceedings of the 2021 Conference of the North American Chapter of the Association for Computational Linguistics: Human Language Technologies}, pages 2736--2746, 2021.

\bibitem{zhang2024cppo}
Han Zhang, Yu~Lei, Lin Gui, Min Yang, Yulan He, Hui Wang, and Ruifeng Xu.
\newblock Cppo: Continual learning for reinforcement learning with human feedback.
\newblock In {\em The Twelfth International Conference on Learning Representations}, 2024.

\bibitem{gao2023unified}
Qiankun Gao, Chen Zhao, Yifan Sun, Teng Xi, Gang Zhang, Bernard Ghanem, and Jian Zhang.
\newblock A unified continual learning framework with general parameter-efficient tuning.
\newblock In {\em Proceedings of the IEEE/CVF International Conference on Computer Vision}, pages 11483--11493, 2023.

\bibitem{peng2024scalable}
Bohao PENG, Zhuotao Tian, Shu Liu, Ming-Chang Yang, and Jiaya Jia.
\newblock Scalable language model with generalized continual learning.
\newblock In {\em ICLR}, 2024.

\bibitem{xu2018reinforced}
Ju~Xu and Zhanxing Zhu.
\newblock Reinforced continual learning.
\newblock {\em Advances in neural information processing systems}, 31, 2018.

\bibitem{aljundi2019gradient}
Rahaf Aljundi, Min Lin, Baptiste Goujaud, and Yoshua Bengio.
\newblock Gradient based sample selection for online continual learning.
\newblock {\em Advances in neural information processing systems}, 32, 2019.

\bibitem{aljundi2019online}
Rahaf Aljundi, Eugene Belilovsky, Tinne Tuytelaars, Laurent Charlin, Massimo Caccia, Min Lin, and Lucas Page-Caccia.
\newblock Online continual learning with maximal interfered retrieval.
\newblock {\em Advances in neural information processing systems}, 32, 2019.

\bibitem{liu2020learning}
Bing Liu.
\newblock Learning on the job: Online lifelong and continual learning.
\newblock In {\em Proceedings of the AAAI conference on artificial intelligence}, volume~34, pages 13544--13549, 2020.

\bibitem{qiao2024gradient}
Jingyang Qiao, Zhizhong Zhang, Xin Tan, Yanyun Qu, Wensheng Zhang, and Yuan Xie.
\newblock Gradient projection for parameter-efficient continual learning.
\newblock {\em arXiv preprint arXiv:2405.13383}, 2024.

\bibitem{bang2022online}
Jihwan Bang, Hyunseo Koh, Seulki Park, Hwanjun Song, Jung-Woo Ha, and Jonghyun Choi.
\newblock Online continual learning on a contaminated data stream with blurry task boundaries.
\newblock In {\em Proceedings of the IEEE/CVF Conference on Computer Vision and Pattern Recognition}, pages 9275--9284, 2022.

\bibitem{aljundi2019task}
Rahaf Aljundi, Klaas Kelchtermans, and Tinne Tuytelaars.
\newblock Task-free continual learning.
\newblock In {\em Proceedings of the IEEE/CVF conference on computer vision and pattern recognition}, pages 11254--11263, 2019.

\bibitem{de2019episodic}
Cyprien de~Masson~D'Autume, Sebastian Ruder, Lingpeng Kong, and Dani Yogatama.
\newblock Episodic memory in lifelong language learning.
\newblock {\em NeurIPS}, 32, 2019.

\bibitem{wang2024clip}
Leyuan Wang, Liuyu Xiang, Yujie Wei, Yunlong Wang, and Zhaofeng He.
\newblock Clip model is an efficient online lifelong learner.
\newblock {\em arXiv preprint arXiv:2405.15155}, 2024.

\bibitem{seo2024just}
Minhyuk Seo, Diganta Misra, Seongwon Cho, Minjae Lee, and Jonghyun Choi.
\newblock Just say the name: Online continual learning with category names only via data generation.
\newblock {\em arXiv preprint arXiv:2403.10853}, 2024.

\bibitem{wang2019symmetric}
Yisen Wang, Xingjun Ma, Zaiyi Chen, Yuan Luo, Jinfeng Yi, and James Bailey.
\newblock Symmetric cross entropy for robust learning with noisy labels.
\newblock In {\em Proceedings of the IEEE/CVF international conference on computer vision}, pages 322--330, 2019.

\bibitem{wei2020combating}
Hongxin Wei, Lei Feng, Xiangyu Chen, and Bo~An.
\newblock Combating noisy labels by agreement: A joint training method with co-regularization.
\newblock In {\em Proceedings of the IEEE/CVF conference on computer vision and pattern recognition}, pages 13726--13735, 2020.

\bibitem{yi2019probabilistic}
Kun Yi and Jianxin Wu.
\newblock Probabilistic end-to-end noise correction for learning with noisy labels.
\newblock In {\em Proceedings of the IEEE/CVF conference on computer vision and pattern recognition}, pages 7017--7025, 2019.

\bibitem{song2019selfie}
Hwanjun Song, Minseok Kim, and Jae-Gil Lee.
\newblock Selfie: Refurbishing unclean samples for robust deep learning.
\newblock In {\em International conference on machine learning}, pages 5907--5915. PMLR, 2019.

\bibitem{han2018co}
Bo~Han, Quanming Yao, Xingrui Yu, Gang Niu, Miao Xu, Weihua Hu, Ivor Tsang, and Masashi Sugiyama.
\newblock Co-teaching: Robust training of deep neural networks with extremely noisy labels.
\newblock {\em Advances in neural information processing systems}, 31, 2018.

\bibitem{yu2019does}
Xingrui Yu, Bo~Han, Jiangchao Yao, Gang Niu, Ivor Tsang, and Masashi Sugiyama.
\newblock How does disagreement help generalization against label corruption?
\newblock In {\em International conference on machine learning}, pages 7164--7173. PMLR, 2019.

\bibitem{jiang2018mentornet}
Lu~Jiang, Zhengyuan Zhou, Thomas Leung, Li-Jia Li, and Li~Fei-Fei.
\newblock Mentornet: Learning data-driven curriculum for very deep neural networks on corrupted labels.
\newblock In {\em International conference on machine learning}, pages 2304--2313. PMLR, 2018.

\bibitem{chen2021beyond}
Pengfei Chen, Junjie Ye, Guangyong Chen, Jingwei Zhao, and Pheng-Ann Heng.
\newblock Beyond class-conditional assumption: A primary attempt to combat instance-dependent label noise.
\newblock In {\em Proceedings of the AAAI Conference on Artificial Intelligence}, volume~35, pages 11442--11450, 2021.

\bibitem{zhang2018generalized}
Zhilu Zhang and Mert Sabuncu.
\newblock Generalized cross entropy loss for training deep neural networks with noisy labels.
\newblock {\em Advances in neural information processing systems}, 31, 2018.

\bibitem{pleiss2020identifying}
Geoff Pleiss, Tianyi Zhang, Ethan Elenberg, and Kilian~Q Weinberger.
\newblock Identifying mislabeled data using the area under the margin ranking.
\newblock {\em Advances in Neural Information Processing Systems}, 33:17044--17056, 2020.

\bibitem{rafailov2023direct}
Rafael Rafailov, Archit Sharma, Eric Mitchell, Christopher~D Manning, Stefano Ermon, and Chelsea Finn.
\newblock Direct preference optimization: Your language model is secretly a reward model.
\newblock {\em Advances in Neural Information Processing Systems}, 36:53728--53741, 2023.

\bibitem{han-etal-2018-Fewrel}
Xu~Han, Hao Zhu, Pengfei Yu, Ziyun Wang, Yuan Yao, Zhiyuan Liu, and Maosong Sun.
\newblock Fewrel: A large-scale supervised few-shot relation classification dataset with state-of-the-art evaluation.
\newblock In Ellen Riloff, David Chiang, Julia Hockenmaier, and Jun{'}ichi Tsujii, editors, {\em Proceedings of the 2018 Conference on Empirical Methods in Natural Language Processing}, pages 4803--4809, Brussels, Belgium, October-November 2018. Association for Computational Linguistics.

\bibitem{zhang-etal-2017-position}
Yuhao Zhang, Victor Zhong, Danqi Chen, Gabor Angeli, and Christopher~D. Manning.
\newblock Position-aware attention and supervised data improve slot filling.
\newblock In Martha Palmer, Rebecca Hwa, and Sebastian Riedel, editors, {\em Proceedings of the 2017 Conference on Empirical Methods in Natural Language Processing}, pages 35--45, Copenhagen, Denmark, September 2017. Association for Computational Linguistics.

\bibitem{li2023online}
Guozheng Li, Peng Wang, Qiqing Luo, Yanhe Liu, and Wenjun Ke.
\newblock Online noisy continual relation learning.
\newblock In {\em Proceedings of the AAAI Conference on Artificial Intelligence}, volume~37, pages 13059--13066, 2023.

\bibitem{li2020dividemix}
Junnan Li, Richard Socher, and Steven~CH Hoi.
\newblock Dividemix: Learning with noisy labels as semi-supervised learning.
\newblock {\em arXiv preprint arXiv:2002.07394}, 2020.

\bibitem{chaudhry2018riemannian}
Arslan Chaudhry, Puneet~K Dokania, Thalaiyasingam Ajanthan, and Philip~HS Torr.
\newblock Riemannian walk for incremental learning: Understanding forgetting and intransigence.
\newblock In {\em Proceedings of the European conference on computer vision (ECCV)}, pages 532--547, 2018.

\bibitem{lopez2017gradient}
David Lopez-Paz and Marc'Aurelio Ranzato.
\newblock Gradient episodic memory for continual learning.
\newblock {\em Advances in neural information processing systems}, 30, 2017.

\bibitem{li2025analyzing}
Xinlong Li, Weijieying Ren, Wei Qin, Lei Wang, Tianxiang Zhao, and Richang Hong.
\newblock Analyzing and reducing catastrophic forgetting in parameter efficient tuning.
\newblock In {\em ICASSP 2025-2025 IEEE International Conference on Acoustics, Speech and Signal Processing (ICASSP)}, pages 1--5. IEEE, 2025.

\bibitem{huang2024mitigating}
Jianheng Huang, Leyang Cui, Ante Wang, Chengyi Yang, Xinting Liao, Linfeng Song, Junfeng Yao, and Jinsong Su.
\newblock Mitigating catastrophic forgetting in large language models with self-synthesized rehearsal.
\newblock {\em arXiv preprint arXiv:2403.01244}, 2024.

\bibitem{touvron2023llama}
Hugo Touvron, Thibaut Lavril, Gautier Izacard, Xavier Martinet, Marie-Anne Lachaux, Timoth{\'e}e Lacroix, Baptiste Rozi{\`e}re, Naman Goyal, Eric Hambro, Faisal Azhar, et~al.
\newblock Llama: Open and efficient foundation language models.
\newblock {\em arXiv preprint arXiv:2302.13971}, 2023.

\end{thebibliography}
